# Detecting value-expressive text posts in Russian social media

Maria Milkova*, Independent Researcher
Maksim Rudnev, University of Waterloo
Lidia Okolskaya, Institute of Sociology FCTAS RAS

*Corresponding author, m.a.milkova@gmail.com

**Abstract**

Basic values are concepts or beliefs which pertain to desirable end-states and transcend specific situations. Studying personal values in social media can illuminate how and why societal values evolve especially when the stimuli-based methods, such as surveys, are inefficient, for instance, in hard-to-reach populations. On the other hand, user-generated content is driven by the massive use of stereotyped, culturally defined speech constructions rather than authentic expressions of personal values. We aimed to find a model that can accurately detect value-expressive posts in Russian social media VKontakte. A training dataset of 5,035 posts was annotated by three experts, 304 crowd-workers and ChatGPT. Crowd-workers and experts showed only moderate agreement in categorizing posts. ChatGPT was more consistent but struggled with spam detection. We applied an ensemble of human- and AI-assisted annotation involving active learning approach, subsequently trained several classification models using embeddings from various pre-trained transformer-based language models. The best performance was achieved with embeddings from a fine-tuned rubert-tiny2 model, yielding high value detection quality (F1 = 0.77, F1-macro = 0.83). This model provides a crucial step to a study of values within and between Russian social media users.

**Acknowledgement**

The authors are grateful to Vladimir Magun for his support and advice during the work on this manuscript.

## INTRODUCTION

Basic values are concepts or beliefs which pertain to desirable end states or behaviors, they transcend specific situations, guide selection or evaluation of behavior and events, and are ordered by relative importance (S. Schwartz, 1992). Values have traditionally been studied by direct prompts to personality (e. g. in a survey), but more recently values in social media content have become a growing area of research interest. Social media reflect societal changes, and studying values in this context can illuminate how and why societal values evolve. Moreover, understanding these values can reveal what unites and divides people on social media platforms.

Research on social media provides a unique source of information about values in populations where stimuli-based methods, such as surveys and experiments, may be limited or inefficient. An example of such a population is Russian society during wartime, where free discussion and expression of "non-traditional" values may be severely punished. Additionally, social media offers an unparalleled resource for investigating values from the past.

Despite increasing interest and a growing number of relevant studies, many approaches were not entirely successful in extracting values. We identified a few potentially problematic points. First, previous studies often failed to distinguish values from other phenomena like personality traits, opinions, or expressions unrelated to personality. Researchers focused on the content topically related to the Schwartz value classification but ignored whether the texts had to do with values as a specific kind of beliefs, possibly leading to biased estimations of value profiles. Second, these studies were designed to describe the values of specific individuals, an approach that could raise ethical and legal concerns in the social media context. User profiling may also be unfeasible since social media posts often follow rigid conversational norms, leaving little room for personality expression. Third, the predominant method for analyzing data, known as the "bag-of-words" approach (or word frequency from a pre-specified list, e.g., Fischer *et al.*, 2022), is limited in its ability to identify complex meanings in text. Consequently, the values extracted this way have shown little correlation with the value profiles obtained through standardized questionnaires (Ponizovskiy *et al.*, 2020; Fischer *et al.*, 2022; Mukta *et al.*, 2016; Chen *et al.*, 2014).

In the current study, we aim to address these shortcomings by proposing an alternative conceptual and methodological approach. The primary goal is to find a model that can accurately detect value-expressive posts, focusing specifically on Russian social media as a critical source of information on population values during times of strict censorship.

## BACKGROUND

### Existing value-extraction studies

Values research has mostly relied on the use of self-report questionnaires (such as Schwartz Value Survey or Portrait Values Questionnaire). However, in recent decades digital user-generated content has also been treated as a behavioral indicator of values. Research interest in studying values through textual data has also been stirring up, as Machine Learning and Natural Language Processing (NLP) methods have quickly evolved.

Although the research on inferring values from texts lags behind similar work on extracting personality traits (e.g., Allport, 1942; Gosling *et al.*, 2003; Catal *et al.,* 2017; Gou *et al.*, 2014, Park *et al.*, 2015), the number of efforts has been growing. The early research explored algorithms that use dictionary-based approaches, the most common of which is Linguistic Inquiry and Word Count (LIWC[1] – Pennebaker *et al*., 2007, 2015). LIWC dictionary includes different thematic categories, which are characterized by their own set of words. LIWC-based analysis assigns each word specific linguistic categories, and presents the total number of words in each category normalized by the total number of words in the document. Early attempts at extracting values were focused on analyzing the association between LIWC categories and individual's self-report scores.

[1] LIWC, URL: https://www.liwc.app

Thus, Chen and colleagues (2014) recruited participants on Reddit, measured their value orientations by PVQ, and used participants' recent posts to identify their score on five value dimensions through the LIWC. The study used correlation analysis and regression models to relate text-based value scores to self-report measures, however found only correlations below 0.2, although unadjusted $R^2$ reached 0.18. They also employed machine learning algorithms to classify individuals with above-median levels of each value dimension but achieved only a small improvement over guessing values on chance.

Boyd with colleagues (2015) asked participants to write value and everyday-behavior essays, conducted topical modeling and calculated frequencies of key topical words. Next, they showed that such measures had a very weak association with a standardized Schwartz Value Survey (SVS). Only three out of 160 unadjusted $R^2$ were greater than .04. Furthermore, using a larger sample from myPersonality application and their Facebook posts, they found even weaker association between topical word frequencies and responses to PVQ. In a similar approach, Mukta and colleagues (2016) recruited Facebook users, offered them PVQ and used three types of users' content: statuses, page-likes, shared links. Linear regression models and classification models were used to select the best subset of LIWC features to represent value dimensions. Using an overfitted model they reached the adjusted $R^2$ of 0.19 for Openness to change values. In further research, this team (Mukta *et al.*, 2018) used Facebook status updates, LIWC and different regression models for temporal value change.

Another study (Silva et al., 2020) recruited 125 undergraduate students who completed SVS and shared their Facebook posts. The researchers used Singaporean modification of LIWC and proposed a new Neural Network model (Stack model) to predict dichotomized personal values. Such a model was able to predict SVS responses with LIWC categories with the AUC ROC between 0.70 and 0.87 however these figures could be inflated given the dichotomization of the criterion variables (MacCallum et al., 2002).

Takayama et al (2015) used an annotated corpus of statements of the Congress hearings on net neutrality and proposed their own method to construct unigrams and bigrams values-dictionary.
Significant step forward in assessing human values through text was the introduction of the Personal Values Dictionary (PVD, Ponizovskiy et al., 2020). The dictionary was specifically designed for the study of values, based on the Schwartz value theory (S.Schwartz, 1992), and has been refined and verified using a variety of sources: Facebook updates, blog posts, essays, and book chapters. Although PVD is just a list of English words divided into 10 value categories (S.Schwartz, 1992), it has stimulated further research on automatic assessment of references to value orientations in textual data. For instance, it has been used to extract values from short autobiographical essays (Fischer *et al.*, 2022) and to analyze texts generated by AI-models (Fischer *et al.*, 2022).

The dictionary-based approach involves the use of a validated theory-driven instrument, and the findings are generalizable, comparable, and replicable (Ponizovskiy *et al.*, 2020). However, the corresponding bag-of-words analyses ignore negations, grammatical structure of texts, poorly recognize semiotic ambiguities, and are prone to misinterpretation (Eichstaedt et al., 2020). On the contrary, data-driven open-vocabulary approaches strongly rely on and are limited to the data specifics. This approach does not use validated sets of words but rely on the lexical composition of the particular text (H.Schwartz *et al.*, 2013). Compared to LIWC (Iacobelli *et al.*, 2011; H.Schwartz *et al.*, 2013; Park *et al.*, 2015) and dictionary-based bag-of-words methods, characteristics of text extracted using the open-vocabulary approach (e.g., words, phrases, and topics) can be better predictors of external variables. For example, Wilson (2019) employed open-vocabulary approach and used topic modeling to discover value concepts in open-ended texts in which people described thoughts about their values, and then applied these models to social media data.

Taking advantage of linguistic information from everyday human language use in social networks can help taking a step forward in studying various psychological constructs (Kulkarni *et al*., 2018), especially when it involves advanced modeling methods such as supervised machine learning based on human coded data (Németh & Koltai, 2021).

Sun et al., 2014 used data from Chinese social networks and proposed Automatic Estimation of Schwartz Values model. To calculate the individual value priorities the model calculated the similarities between a word vector of the personal posts and each value feature vector (that was previously constructed based on detailed values defined by Schwartz (2012), their synonyms, and most common words from news related to specific value category).

Lin et al., 2017 used task-specific Twitter data (with hashtags relevant to Hurricane Sandy), utilized entity linking to external knowledge base (Wikipedia), and incorporated background knowledge to improve moral values detection.

Mukta et al. (2021) predicted the value change of Facebook users (selected randomly but volunteered their data to the researchers) by considering the influence of friends and social media usage. Authors extracted ego networks, segmented users' interaction in a 6-months frame, computed users' value scores via IBM Watson API, and used Bounded Confidence Model for value dynamics analysis.

A separate body of research focused on the automatic extraction of values from public debates (Fleischmann et al., 2009, 2015) and e-mails (Zhou *et al.*, 2010). Qiu *et al.* (2022) used social scenarios with value-related keywords, annotated them via Amazon Mechanical Turk, and trained deep learning models to developed value-driven chatbot.

The most recent studies focused on the task of identifying the values within textual arguments. The data for the research was collected primarily for the task of constructing argumentation on socially relevant topics (Mirzakhmedova *et al.,* 2023). Thus, the best models (Schroter *et al.,* 2023) achieve high accuracy only when faced with a text containing a socially relevant topic on which an argument is given. Importantly, the models do not show an adequate result if the text does not reflect any value at all.

In short, recent advances in automatic value extraction are limited by the specificity of the data used, and work with texts that have been selected according to often arbitrary criteria. This significantly constrains the generalizability of the models. The best results are achieved when texts with social scenarios or discussions of socially relevant topics are used.

Advances in the development of generative Large Language Models (LLMs) have made it possible to introduce knowledge about the expression of values in the development of specialized dialogue systems. Researchers realized that communication requires an understanding of values and norms (Forbes et al., 2020). It is an uncharted territory and still needs caution and human check, given its non-transparency and the fact that minor wording alterations in prompts or repeating the identical input can lead to varying outputs (Reiss, 2023). Nevertheless, there is early evidence that ChatGPT in particular outperformed crowd-workers in annotation task (e.g., Törnberg, 2023; Gilardi et al., 2023).

Finally, and critically, most research, both dictionary- and text-based, focused on English language (Dobbrick *et al.,* 2022) and cannot be directly transferred to non-English textual data and therefore such studies are scarce (for exceptions, see Sun *et al.*, 2014; Khayrullin *et al.*, 2017; Bogdanov, 2020; Kalabikhina *et al.,* 2023). For the dictionary-based approaches, the problem is that direct translation to another language may bring bias and increase the error. Indeed, dictionaries designed for a specific language usually work better than the translated ones (Panicheva & Litvinova, 2020). For the text-based approaches, the supervised classification models require data annotations for a specific use case in the target language (Chan *et al.*, 2021).

## CURRENT STUDY

User-generated content in social media is driven by the massive use of stereotyped, culturally defined and desirable speech constructions rather than tendency to express one's own values. Indeed, self-representation in social media is mediated by audience expectations (Marwick *et al.*, 2011), as well as by the level of freedom of speech and regulations in a country and rules of a particular social network. Thus, an important and previously unexplored venue of research is the analysis of value climate in social network, which implies, first, going beyond the personality of the author, and second, distinguishing value-expressive content.

### "The author is dead"

In our approach, we do not make a connection between the text and its author but assess texts in terms of its perception by a native perceiver. Following from the prominent semiotic theory developed by Roland Barthes (1977), we claim that the authorship (author's intentions, motivations, personality) does not matter when the central interest is interpretation of text itself. In this approach the focus is moved towards *a reader* or a consumer of social media posts. The focus on the reader is necessary when we aim to describe values of the entire online community because the vast majority of content is generated by a very small proportion of users (Schneider *et al.*, 2013). Therefore, profiling of users based on their posts would reflect values of a small and atypical group. Moreover, a substantial share of posts is created by bots today (Ferrara *et al.*, 2016), and this share is likely to be growing following the hyperbolic development of generative AI.

Finally, trying to identify user's values based on their social media posts might be a futile quest because the spontaneous non-prompted verbal behavior might contain too much noise. Perhaps this is why previous attempts found at best a weak association between an author's personal values and value expressions in their writings. Indeed, the verbal (posting) behavior is more likely to follow ritualized norms, aim at image maintenance, and other communication-related goals rather than directly express values of a user.

Therefore, our focus is moved towards perceivers and the value context they exist in their online lives, or, more specifically, on value profiling of distinct posts.

By value-expressive content we mean posts that explicitly or implicitly express the importance of any idea, phenomenon, or quality. Value-expressive texts are different from personal values in that the value appears when the social media user reads the post (Barthes, 1977). By reading the value-expressive post the user's values are primed. Intention to read the new updates (by the means of subscribing to/following/friending) a user might show their agreement with the expressed values (although the extent to which it happens may vary).

### Distinguishing value-expressive posts

Determination of whether a post is value-expressive or not is a necessary first step before the post can get its value profile. Although missing from the previous studies, which did not attempt to differentiate between values and other subjective phenomena such as norms or opinions, determining whether the post is value-expressive is important for the following reason. As mentioned above, the texts in social media can follow ritualized expressions and are often led by the social norms which do not necessarily carry any value content. Focus on the themes described in Schwartz classification of the value content while skipping the stage of detecting whether the post expresses a value, might lead to a biased picture of a value climate. Moreover, in restricted situations, such as censorship, the immediate meaning of a post might become elusive and ambiguous, which requires more fine-tuned tools, such as extension of the research with one preliminary step of identifying the very fact of value expression. Therefore, it is important to distinguish between the posts that look like values content-wise (e.g., emphasizes benevolence but does not point to its importance or is used in a ritualized way, by saying "have a good weekend") and posts that do express values as perceived by the social network users. The current study is the first step towards building a classification model that will distinguish between specific values, whereas the purpose of this step is to develop a model that is capable of distinguishing between posts which express values and those that do not.

## DATA AND METHODS

### Data collection

This work uses the VKontakte social network, which is the most used social network for Russian-speaking audience[2]. To ensure representation of all VKontakte users, we have selected random users by randomly generating 30,000 id numbers (from 1 to 710000000). The posts were collected between August 5 and August 26, 2022. We used VKontakte API to access the posts. The analysis was carried out in the Python environment.

Only 5,621 (18.7%) of the initially generated IDs had an open profile and had at least one text post on the wall. We collected all posts from the users' feeds ("walls") and included only text posts written in Cyrillic, consisting of at least one two-word phrase – overall 129,437 posts (out of 188,246 initially collected text posts).

### Overall study design

The aim of our study was to develop a model capable of detecting value-expressive social media posts. Given that the effectiveness of classification algorithms is closely tied to the quality of the training and validation data (Song *et al.*, 2020), we adopted a thorough method to acquire criterion data. Therefore, the study focused on accuracy, reliability, and bias across annotation methods before we proceeded to develop the actual classification model. Figure 1 shows the general flow of the data preparation and annotation.

First, we filtered the data by detecting spam and rubbish posts. We subsequently allocated trial and training subsets from the main dataset. Second, we utilized three different approaches to annotate social media posts from the trial dataset (see below). The trial subset included 804 randomly selected posts. This number was determined mostly by budgetary restrictions. These were utilized to test various annotation methods and later served as high-quality material for validating automated text classification procedures. Third, we assembled a training dataset (5,035 posts), further annotating posts with uncertain classifications. Finally, we trained several language models with different word embedding techniques to identify the best one.

All study materials, including annotation guideline, VKontakte API scripts, scripts for data preprocessing, annotation, and classification are available in the GitHub repository (https://github.com/mmilkova/nlp-values).

[2] According to data provided by Similarweb. URL: https://www.similarweb.com/ru/website/vk.com/#geography

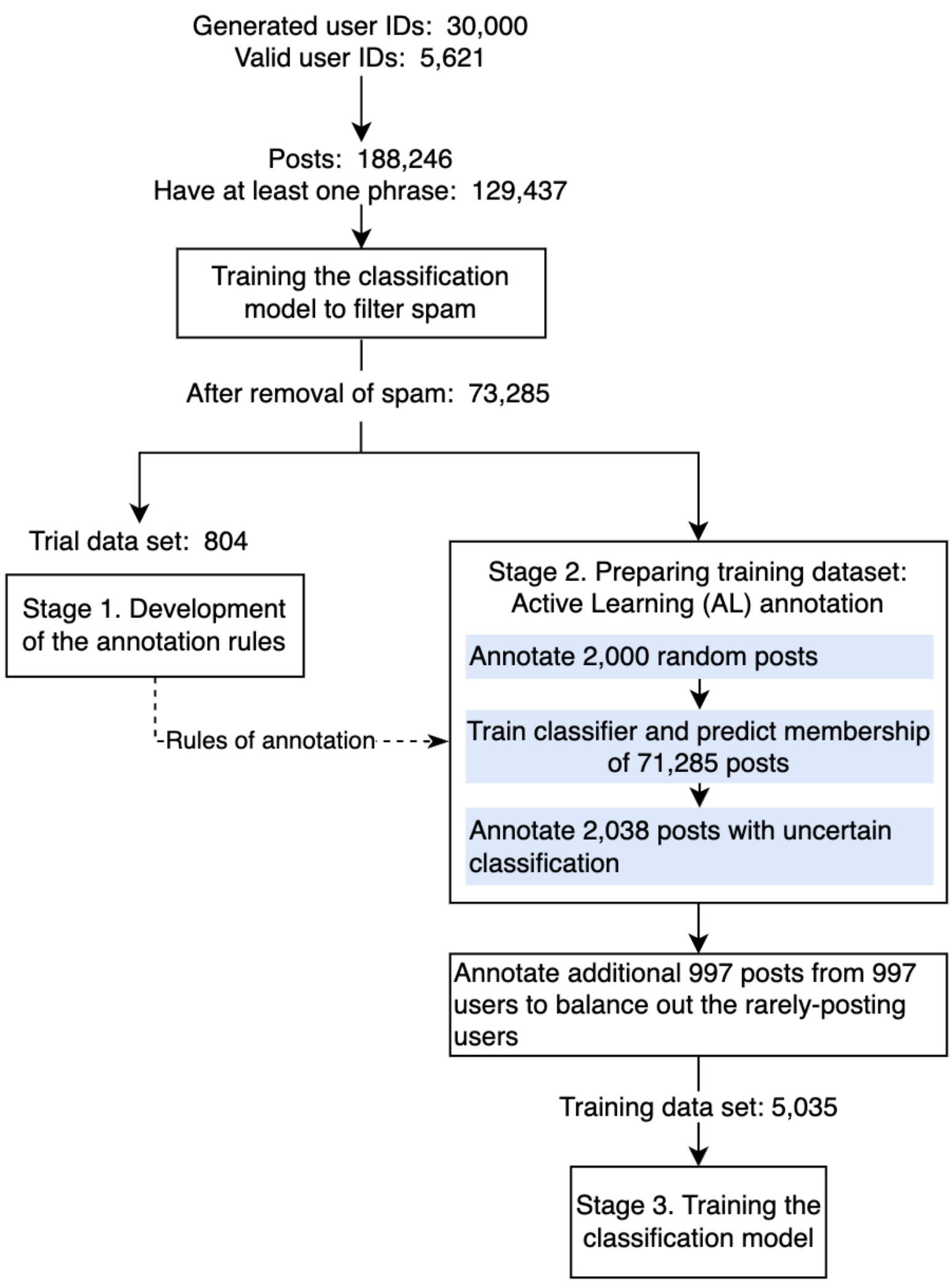

Figure 1. Overview of data samples and studies

## Spam detection and filtering

The VKontakte data was noisy, containing a significant number of spam posts, such as advertisements and auto-generated messages from VKontakte applications. While spam and commercials might influence perceivers and shape the "value climate", they are often viewed with skepticism as long as social media users recognize that these posts aren't genuine personal content. As such, we opted to exclude spam/commercial posts from further analysis. Manually filtering this data would be time-intensive, so we turned to topic modeling to compile a set of key phrases, subsequently setting simple rules to flag spam. Through weakly supervised learning classification, we determined that 32% of the messages were spam. Beyond filtering spam, we pre-processed the data to include only texts written in Russian, removed user IDs, private information (such as phone numbers, emails, and card numbers), and eliminated duplicate posts. After these procedures, the dataset consisted of 73,285 posts.

## Annotators

As mentioned above, our study is focused on the perceived values. Thereby the annotations should be produced by the individuals familiar with the cultural and platform-specific context, and have a good sense of the concept of values: we considered both human coders (experts and crowd-workers) and artificial intelligence (AI).

**The first** group of annotators was comprised by three experts, specialists in the field (the authors). Expert annotation was assumed to be the gold standard and implied that expert understanding of values coincided with its specific understanding (that is, corresponds to Schwartz definition).

**The second** group of annotators was crowd-workers. We recruited Russian-speaking coders through the crowdsourcing platform Yandex Toloka (equivalent of Amazon Mechanical Turk, see - https://toloka.ai). In this case, we followed an ethnographic approach – that is, we considered the perception of VKontakte posts by the natives, VKontakte users themselves (see Appendix 1 for more details).

Overall, 90 crowd-workers participated in the annotation of the trial dataset, 36 of them were banned (30 failed control tasks, 6 were banned because of too fast responses). Thus, annotations from 54 crowd-workers were accepted, each of which annotated on average 50 posts (median is 35, the minimum value is 10, the maximum is 212). For more details on the annotation process, see Appendix 1.

**The third** annotation approach employed ChatGPT. There is early evidence of its cost-effectiveness in text annotation (see e.g., Gilardi et al., 2023), moreover, ChatGPT has already demonstrated promising results in identifying value content (Fischer et al., 2023). Nevertheless, it also caused some concerns due to the non-transparency of the technology (Reiss, 2023). Thereby, we used it as an additional source of annotation.

We used the gpt-3.5-turbo model via the OpenAI API with temperature=0 which implies the least randomness and leads to higher consistency and more deterministic responses[3]. We used the same annotation guide as with the other annotators. As explained below, the guide had several examples of posts and instructions on how they should be classified, we can assume that we used ChatGPT with a few-shot learning (i.e., we show the model several examples of proper annotation).

Each labeling approach provided three annotations for each post. To measure the intercoder consistency (sometimes referred to as reliability), we used Krippendorff's Alpha (Krippendorff, 2004). Since the reliability of the majority label is often higher than any individual label (Törnberg, 2023), we used the majority labels for comparing different labeling strategies (that is, when two out of three crowd-workers provided the same annotation). In cases when all the coders assigned different labels, the post was not assigned any rating and marked "Not assigned". We measured annotation accuracy as the proportion of crowd-worker and ChatGPT annotations that agree with the final expert classifications. We also calculated Precision (P), Recall (R) and F1 metrics.

Since the purpose of this study is the detection of value-expressive posts, we aimed to minimize false negative results (i.e., when the post is value-expressive, but was misclassified and therefore lost for further study of value content). To do that, we conducted the bias analysis that compared classifications produced by different annotators using confusion matrices.

### Annotation guide

In order to ensure that the annotators detect value-expressive posts and not the adjacent phenomena such as norms or traits, we developed a simplified but very accessible definition of values which is close to the formal definition given by Schwartz (1992). Among many possible options we opted for the following definition of value-expressive content as "the importance of some idea, phenomenon or quality". This definition is accessible, is clear to non-specialists, it emphasizes "importance" as a distinguishing feature of values, which is often used in survey measurement instruments of values, and is broad enough by mentioning ideas, phenomena, and qualities.

[3] OpenAI documentation, URL: https://platform.openai.com/docs/models/gpt-3-5

We have developed a single guideline that explained what basic human values are, and provided real examples of posts that express and do not express values.

Experts, crowd-workers, and ChatGPT were instructed to follow the same guidelines. All of them answered the question: "*Does this post reflect the importance of some idea, phenomenon, or quality?*" and classified each post to one of the following categories: *"Reflects values", "Doesn't reflect values", "Unclear", or "Spam"*. The last option was added as a few spam posts still remained in the dataset. The full wording of the guideline is provided in Appendix 1.

**Active learning approach**

Active Learning (AL) strategy is a technique of sequential annotation. Instead of selecting posts for annotation randomly, it selects posts based on their usefulness for the model (Settles & Craven, 2008; Settles, 2009). According to AL, selection of posts for annotation requires assessment of their informativeness for the model. The most common measure of informativeness is the degree of posts' classification uncertainty: higher uncertainty implies the lack of information which could be added by the annotation. Thus, AL implies that the posts for annotation are selected iteratively, based on the results of modeling. This approach is widely used to solve many NLP problems, helps to reduce the amount of annotation required, and has already shown its effectiveness (Siddhant and Lipton, 2018; Rogers *et al.,* 2018; Shelmanov *et al.,* 2019). We applied it at the stage when the training dataset was formed.

**Classifier models**

Supervised Machine Learning often uses manual coding to annotate datasets. This process involves human annotators labeling the data based on predefined categories or criteria. When training the classifier, not the words themselves are taken as initial data, but their vector representations, called word embeddings. This approach avoids the problem of synonymy and homonymy, since words with similar meaning will have the same representation. Trained ML model was supposed to automatically predict the four annotation categories using the text data.

**RESULTS**

**Annotations**

Table 1 presents the frequency distributions and the reliability and accuracy measures for all annotation approaches on the trial dataset of 804 posts. Although the understanding of the values by the experts was expected to be similar, consistency between them was moderate – Krippendorff's Alpha is 0.54 (and 0.57 when merging "Doesn't reflect" and 'Unclear' classes). In a crowd-worker annotation, intercoder[4] consistency was 0.44 and 0.46, respectively. Consistency seems to depend on the content of the post: eye-balling of posts with different consistency has shown that posts with sentiments and emotions (declarations of love, reflections on happiness, mental anguish) have higher consistency. Accuracy of crowd-workers annotation was 0.71. P, R and F1 for the 'Reflects' class were 0.64, 0.65, and 0.65.

ChatGPT had much higher internal consistency (0.95/0.98) compared to human annotators. Accuracy was 0.57, but merging 'Doesn't reflect' and 'Unclear' classes increased it significantly up to 0.72. P, R and F1 for the 'Reflects' class were 0.56, 0.76, and 0.65.

Frequency distributions across response classes (the majority classes) are close for experts and crowd workers, whereas ChatGPT tended to label posts as 'Reflects' or 'Unclear', and fewer posts were classified as Spam.

[4] Note that although three codes always came from three different annotators, the annotators could differ for different posts. For example, post A could be annotated by coders #1, #2, #3 and post B by annotators #2, #25, #31.

Table 1 – Frequency distributions of classes for three annotation approaches and Reliability and accuracy of labeling by crowd-workers and ChatGPT on test data (N = 804)

| | Experts | Crowd-workers | GPT 3.5 | Crowd-workers + GPT combined using the rule |
|---|---|---|---|---|
| Reflects | 29.7 | 30.0 | 40.4 | 33.8 |
| Doesn't Reflect | 58.7 | 57.7 | 37.7 | 55.8 |
| Unclear | 0.0* | 0.5 | 19.0 | |
| Spam | 9.0 | 9.8 | 2.9 | 10.3 |
| Not assigned | 2.6 | 1.9 | 0.0 | 0.0 |
| Krippendorff's Alpha | 0.54 | 0.44 | 0.95 | - |
| Krippendorff's Alpha (merged 'doesn't reflect' and 'unclear') | 0.57 | 0.46 | 0.98 | - |
| Accuracy | 1 | 0.73 | 0.57 | - |
| Accuracy (merged 'doesn't reflect' and 'unclear') | 1 | 0.74 | 0.72 | 0.76 |
| P/R/F1 (for 'reflects' class) | 1/1/1 | 0.64/0.65/0.65 | 0.56/0.76/0.65 | 0.64/0.73/0.68 |

Accuracy is a proportion of value-expressive posts classified in agreement with expert labels. When calculating the accuracy, 'Not assigned' majority class was combined with the class "Doesn't reflect'.

* Experts annotated posts without the "Unclear" category, but it was added for crowd-workers and GPT.

For the bias analysis and ease of use of the model in the future, we combined the classes 'Unclear' and 'Doesn't reflect'. Figure 2 shows confusion matrices: the percentage of each of the classes annotated by crowd-workers (A) and ChatGPT (B) against the classification given by the experts, whereas panel C shows the percentage of each of the classes annotated by crowd-workers versus classification given by ChatGPT.

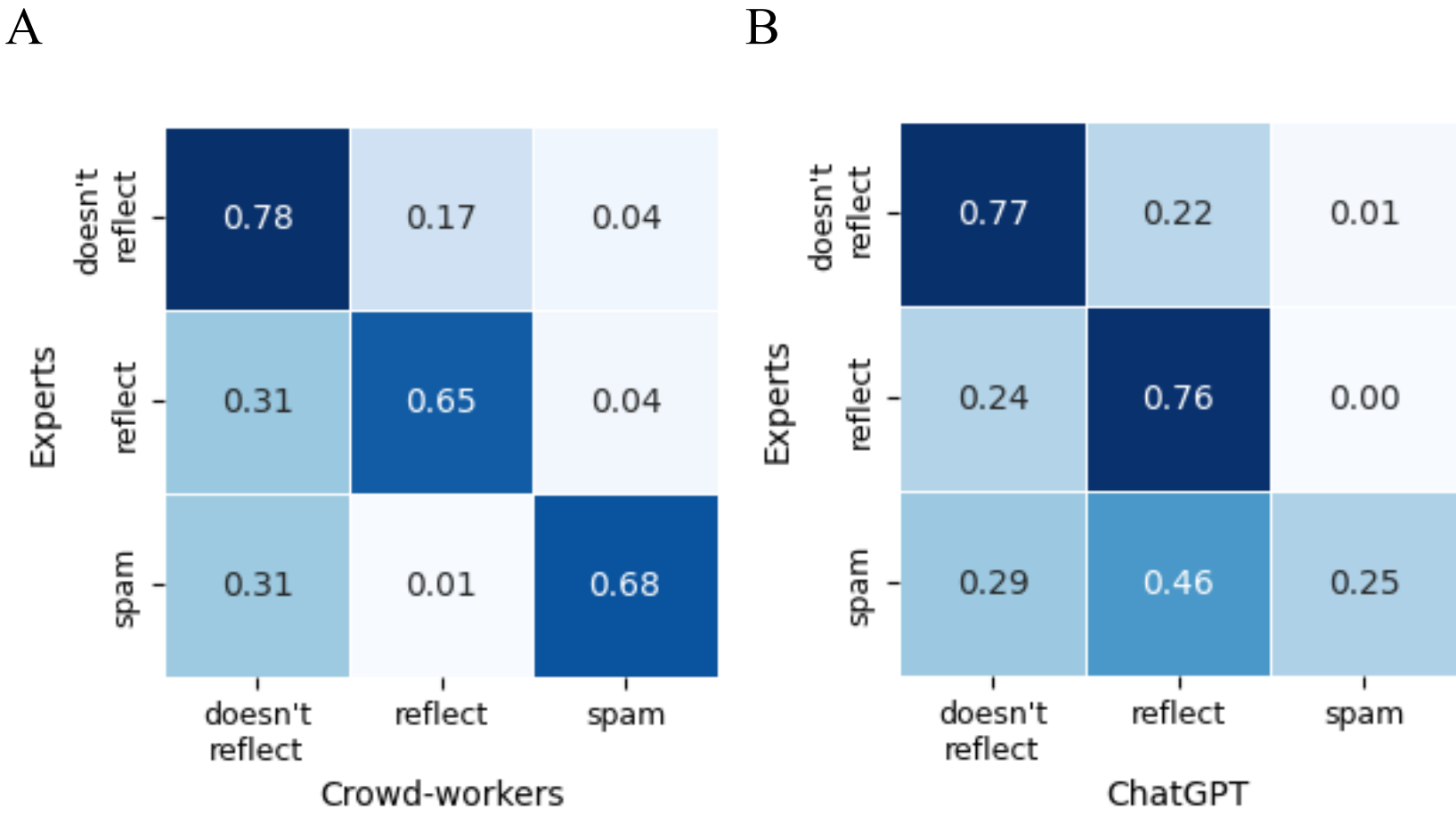

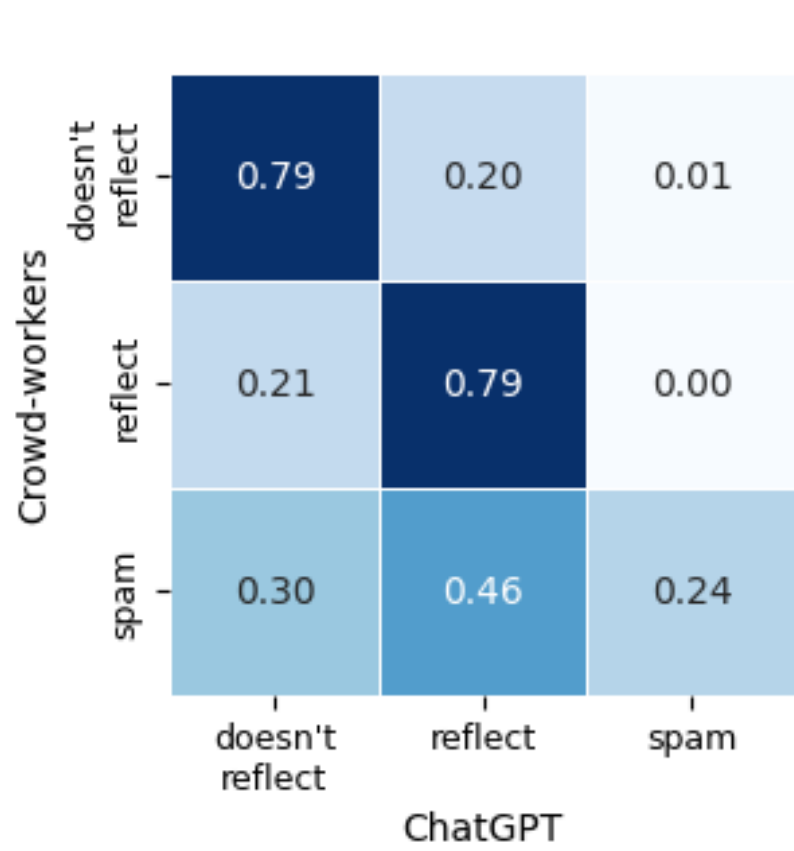

Figure 2. Confusion matrices for experts and crowd-workers (a), experts and ChatGPT (b), crowd-workers and GPT 3.5, percent by row.

Only 65% of posts classified by experts as value-expressive were classified in the same way by crowd-workers. Examination of misclassified posts[5] showed that the crowd-workers failed to recognize values in three categories: political issues (e.g., "Do the ministers of economy and finance offer other options besides retirees? Or is this their only and last strategic reserve?"), posts about parents being proud of their children (e.g., "Another one of my princesses has grown up!"), and posts about traveling or enjoying life (e.g., "Well, do not feed bread, let me go somewhere! ['do not feed bread, let me do smth.' - expression in the Russian language, means a strong desire to do smth.]", "It's cool to snooze on the oceanfront wrapped in a towel🌊 # dominican#ocean#atlantic#ocean#palms🌴#vacation").

There were 17% of false positive errors, that is, when crowd-workers mistakenly classified post as value-expressive and experts did not. Among these were the posts containing sentiments, "folk wisdom", grandiloquent statements, and posts about romantic relationships.

The percentage of value-expressive posts that were accurately classified by ChatGPT was 76%. Examining cases where ChatGPT labeling was false positive (22% of the posts that experts and 20% of the posts that crowd-workers classified as not value-expressive) highlighted the presence of words that might have indicated value-expressive content, like "love", "courage", and "reward", but the overall meaning of the posts did not seem to express any value. Here are few examples of such posts:

[5] Here and below, we provide English translation of the posts. The original texts can be found in Appendix 5.

> There are no accidents - everything in this world is either a trial, a punishment, a reward, or a foreshadowing. ©Voltaire
>
> If a guy loves you, he'll never let other people hurt you. He'll hurt you on his own.
>
> Instruction before the path of courage))
>
> – Honey, at least start running in the morning... – If I start running, I will not come back!!!

False negative results of ChatGPT were also related to posts with implicit value-expression (e. g., *"*Sometimes, I want to go far away from everyone, to a faraway place, not to see or even hear anyone.*",* "The best therapist is a friend with a bottle of champagne."), and short incomplete posts (e.g., "Friends are everything to me (c)", "I'm finally an auntie!!!!!!!!))))))"). In contrast to ChatGPT, posts that imply the completion of meaning were successfully classified by crowd-workers. All in all, ChatGPT failed in cases where classification required particular interpretive skills.

Moreover, ChatGPT very often misclassified spam: 'native' ads, asking for help postings, promo for business development. It misclassified 75% of spam identified by experts and 76% by crowd-workers. This highlights the fact that many promotional posts were often accompanied by value-expressive content (self-improvement announcements, fundraising posts, etc.). Overall, ChatGPT tended to provide more false positive answers, whereas crowd-workers tended to give false negatives. For instance: "Girls! 🙋🏻‍♀️ What do you think the perfect makeup bag should consist of? 🤨 Of course, of those products that are just necessary to look perfect ☝ Cosmetics [Brand name removed] ❌ is just created for the perfect beauty kit. <…>"

### *The final annotation rules*

We decided to combine the crowd and ChatGPT results to get annotations with higher accuracy. We primarily relied on the classifications provided by ChatGPT, but aimed to reduce the number of ChatGPT misclassifications by incorporating the level of interrater agreement for each post. The goal here was to replace the ChatGPT classes referring to "Doesn't reflect", "Reflects" when all crowd workers unanimously assigned the post to a different class. The exception was the Spam class. As the previous section showed, crowd-workers were much better than ChatGPT in detecting spam. Therefore, in the final combined annotation rules we classified posts as spam if they had been classified as spam by crowd-workers. 'Not assigned' class was combined with "Doesn't reflect'.

Thus, for the final annotation we used classification provided by ChatGPT, overriding it only when the majority of crowd-workers identified post as 'Spam' or 'Reflects' or when all of them classified the post as "Doesn't reflect". This strategy revealed the accuracy of 0.76 (see right column of Table 1 for frequencies) and P/R/F1 for the 'Reflects' class of 0.64/0.73/0.68.

Confusion matrix between the final and expert results is shown in Figure 3. Thus, despite the fact that the annotation of both crowd-workers and ChatGPT was characterized by average accuracy, combining the two labeling approaches improved the control over missing value-expressive posts, and better detected posts that are inexpressive of values.

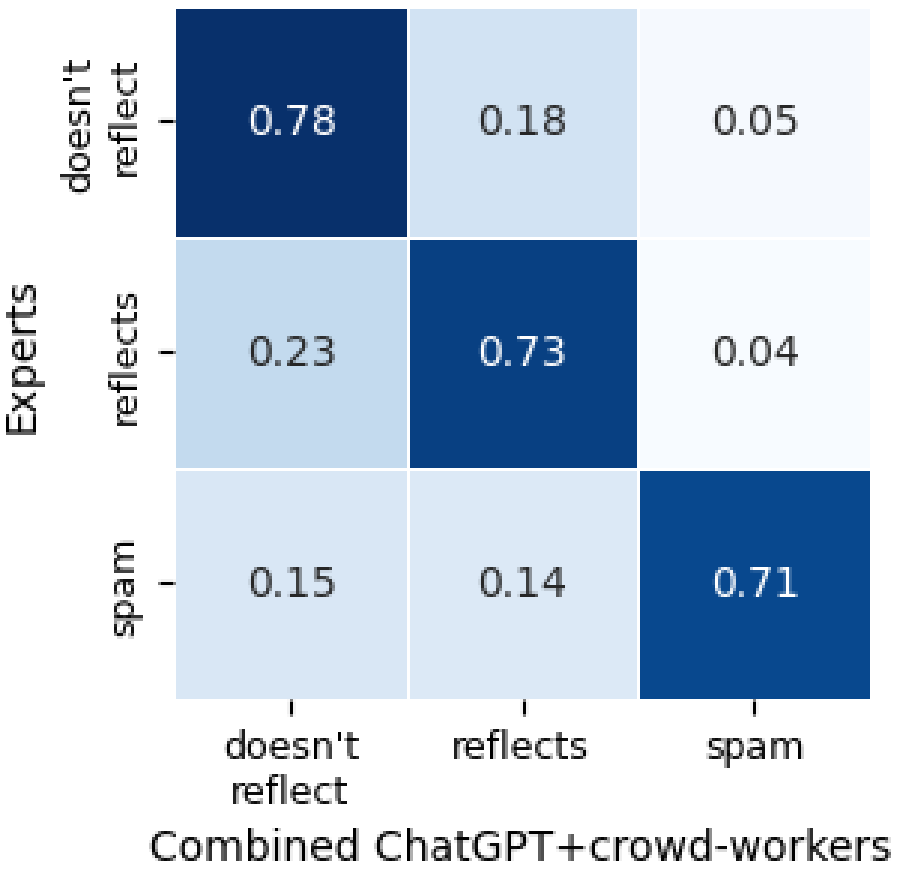

Figure 3. Confusion matrix (percent by column)

**Forming the training dataset**

Next, we used the new data (training dataset) and applied the rules of annotation[6] developed in the previous section. We conducted the AL approach to be as follows:

1) From the main dataset of 73,285 posts, we randomly selected 2,000 posts and annotated them with help of crowd-workers and ChatGPT.
2) Next, using these annotations, we trained a classifier model, namely Support Vector Machine (SVM) model based on embeddings from the pre-trained *rubert-tiny2*, since in preliminary experiments it yields better computational performance[7].
3) The trained classifier model subsequently predicted the probability of classes for the remaining 71,285 posts.
4) Next, we filtered out the posts that had high uncertainty of classification (probability in the range from 0.3 to 0.7), randomly selected 2038 posts from this sample, and annotated them. Following this strategy, we passed new information to the model.

As an additional step, we annotated yet another 997 posts. Social network users different tremendously by the number of posts created. To balance out the representation of a larger number of users in annotation, we selected 997 users and picked one random post from each, and annotated it using crowd-workers and ChatGPT.

As a result, our training dataset consisted of 2000+2038+997=5035 posts that were annotated by 304 crowd-workers, ChatGPT, and their annotations were combined following the rule described above. The distribution of annotations (see Table 4 for distribution and Appendix 1 for more details) was similar across the trial and training datasets: ChatGPT tended to misclassify *Spam* posts, consequently, the percentage of value-expressive posts was higher than the one among crowd-workers (44.8% and 30.6% respectively). ChatGPT also tended to label posts as *unclear* more often (19.2%,) than crowd-workers (0.4%). Consistency remained moderate among crowd-workers ($\alpha$ = 0.39) and high ($\alpha$ = 0.96) between ChatGPT labeling attempts.

[6] The sample size was determined by the budget of the study and some technical considerations related to cleaning the data from noise.
[7] Small distilled version of the bert-base-multilingual-case model for Russian and English. URL: https://huggingface.co/cointegrated/rubert-tiny2

Table 4 – Distribution of annotations in the training dataset, per cent (N = 5,035)

| | Crowd-workers, $\alpha$ =0.39 | ChatGPT 3.5, $\alpha$ = 0.96 | Final Label |
|---|---|---|---|
| Reflects | 30.6 | 44.8 | 34.4 |
| Doesn't Reflect | 54.9 | 34.2 | 54.4 |
| Unclear | 0.4 | 19.2 | |
| Spam | 10.7 | 1.8 | 11.1 |
| Not assigned* | 3.5 | 0.0 | 0 |

* all three annotations were different

## Classification models

We used our training dataset (N = 5,035) with the final labels for development of the target classification model. We aimed to conduct binary classification (value-expressive vs. not value-expressive) and at this stage we considered all spam posts as value-inexpressive, and thus combined the 'Spam' class with the 'Doesn't reflect class'.[8]

Quality of the classification models was evaluated by calculating Precision, Recall, and F1 scores on the trial dataset, against the final labels obtained in the previous section. We chose to use these metrics because the classes are unbalanced ('Doesn't reflect' class prevails over "Reflects" class). The target evaluation metric is F1 for the "Reflects" class. We also provide F1-macro score as an unweighted mean of F1 metrics for both "Reflects" and "Doesn't reflect" classes.

We applied 5-fold cross-validation and experimented with the following models. First, we trained classical models (LogitRegression, LogitBoost, SVM) based on pre-trained word-embeddings (768-embeddings from Conversational RuBERT[9], 1024-embeddings from SBERT[10]). LogitRegression, LogitBoost, and SVM classical models were used because they have shown to perform well in preliminary experiments.

Second, we fine-tuned rubert-tiny2[11] for five epochs, extracted the 312-embeddings and fed them to SVM, LogitBoost and LogitRegression models. For each of the models we applied Grid search to find the best model parameters.

We chose Conversational RuBERT, because it was pre-trained on social media data, and SBERT model – as it is a large multi-task model. We chose rubert-tiny2, because as it is a distilled small model, that can be quickly fine-tuned. Previous findings have shown that distilled models yield better computational performance and reduce obstacles for applying deep learning (Shelmanov *et al.,* 2021). All the models are focused on the Russian language.

The performance of classification is listed in Table 5. In terms of F1 for "Reflects" class, the highest score is achieved with the SVM model based on embeddings, extracted from fine-tuned rubert-tiny2 model and resulting in 0.768. F1-macro is 0.826.

SVM model using embeddings from the fine-tuned rubert-tiny2 model achieves the highest score of 0.768 for the "Reflects" class. Its F1-macro score is 0.826. Overall, the best results were shown by models based on embeddings from fine-tuned rubert-tiny2, which were previously pre-trained, including on data from social media. Thus, in our experiments the quality of a model is determined not by its size, but rather by the data on which the models were pretrained.

[8] In future research, we plan to include 'Spam' class to refine classification models.
[9] Conversational RuBERT – 178M parameters model. URL: https://huggingface.co/DeepPavlov/rubert-base-cased-conversational
[10] SBERT – 427M parameters model. URL: https://huggingface.co/ai-forever/sbert_large_mt_nlu_ru
[11] Rubert-tiny2 – 29M parameters model, distilled version of BERT multilingual base model for Russian and English. URL: https://huggingface.co/cointegrated/rubert-tiny2

Further, we used the SVM Fine-tuned rubert-tiny2 model to classify all the posts in our main dataset (73,285 posts), showing that 32% of them (23,300) were value-expressive posts. Given the noisy data, this proportion is higher than could be expected but given several stages or preliminary data cleaning, seems plausible.

Table 5 – Binary classification results with embeddings from deep learning models

| Model and embeddings | Class 0: "Doesn't reflect" | | | Class1: "Reflect" | | | F1-macro |
|---|---|---|---|---|---|---|---|
| | Precision | Recall | F1 | Precision | Recall | F1 | |
| *SVM* | | | | | | | |
| Conversational RuBERT | 0.834 | 0.914 | 0.872 | 0.792 | 0.643 | 0.710 | 0.791 |
| SBERT | 0.844 | 0.890 | 0.870 | 0.770 | 0.676 | 0.720 | 0.794 |
| Fine-tuned rubert-tiny2 | 0.875 | 0.895 | 0.885 | 0.785 | 0.750 | 0.767 | 0.826 |
| *LogitBoost* | | | | | | | |
| Conversational RuBERT | 0.821 | 0.895 | 0.856 | 0.750 | 0.618 | 0.674 | 0.767 |
| SBERT | 0.841 | 0.876 | 0.858 | 0.736 | 0.676 | 0.705 | 0.782 |
| Fine-tuned rubert-tiny2 | 0.874 | 0.874 | 0.874 | 0.753 | 0.753 | 0.753 | 0.814 |
| *LogitRegression* | | | | | | | |
| Conversational RuBERT | 0.855 | 0.921 | 0.887 | 0.818 | 0.695 | 0.751 | 0.819 |
| SBERT | 0.864 | 0.863 | 0.864 | 0.733 | 0.735 | 0.734 | 0.799 |
| Fine-tuned rubert-tiny2 | 0.880 | 0.876 | 0.873 | 0.746 | 0.768 | 0.757 | 0.815 |

## DISCUSSION AND CONCLUSION

The aim of this study was to create a model that can predict whether the post in Russian social media is value-expressive or not. We scraped an initial sample of posts, cleaned it, and explored three different ways to annotate the data.

First, we observed that human coders, both crowd-workers and experts, showed only a moderate agreement on the attribution of value-expressiveness to the posts. This observation can be explained by the fact that humans bring their own understanding of what values are, which leads to a variation of interpretations. This finding is consistent with other studies, which showed that experts, while having conceptual knowledge of the subject, do not provide greater consistency than untrained coders (Weber *et al.*, 2018). On the one hand, it is something to be expected (Aroyo and Welty, 2015; Weber *et al.*, 2018). On the other hand, it causes higher uncertainty in the data entered in the models which in turn may reduce its predictive power. This uncertainty can be overcome by a larger size of annotated dataset (Song *et al.*, 2020), and future research should maximize it.

Beyond the moderate agreement within groups of experts and crowd-workers, we found systematic divergence between experts and crowd-workers. Interestingly, crowd-workers were reluctant to classify posts on political issues, parental pride, and enjoyment of life as value-expressive. While this observation requires further testing on a larger dataset, it raises the question of how the annotation results are correlated with characteristics of the coder. While Weber *et al.* (2018) argued that extraction of moral intuition from texts can be predicted by a coder's political attitudes, it can be assumed that a value profile of a crowd-worker also affects their annotations. Thus, in order to avoid bias in human coding, it may be appropriate to identify a group of coders with a specific profile that will meet the annotation task for texts of different complexity.

Second, following the recent advances in artificial intelligence, we experimented with ChatGPT for annotation of posts. ChatGPT has a high level of consistency, and that is directly dependent on *temperature* setting. In our study we only experimented with *temperature* set to zero, as it seemed unnecessary to add creativity and riskiness to the answers.

The accuracy of recognizing value-expressive posts varied widely between crowd-workers and ChatGPT. Most notably, ChatGPT failed to recognize spam, ads, and other promotional posts, and marked them as value-expressive. Although some of the spam posts are technically value-expressive, they should not be considered as such because their perception deviates from authentic postings by humans. Crowd-workers were much more accurate in recognizing spam. Another reason may come from the fact that "value-expressiveness" and "spam" are not mutually exclusive categories and it brings certain ambiguity, and crowd-workers as primary consumers of the social media information do recognize that "spam" should override "value-expressiveness" whereas ChatGPT does not. Additional testing or using later versions of ChatGPT might provide insight into how to mitigate this kind of bias.[12]

On the other hand, ChatGPT was more accurate in recognizing values in posts where the crowd-workers failed (e.g., political issues). In the future, it will likely be possible to completely replace crowd-workers with ChatGPT in the text-annotation task. However, according to the current state of affairs, human coding is still necessary, since ChatGPT requires validating its performance for each specific task.

Altogether, the non-overlapping errors of crowd-workers and ChatGPT allowed us to combine the annotations from both crowd-workers and ChatGPT, which compensated for each coder's errors. Next, we employed the Active Learning technique that ensured that only ambiguous posts were annotated and helped to reduce the burden of annotation. Taken together, this created an efficient and cost-effective pipeline of value posts annotation.

Using the resulting data for training different kinds of machine learning models, we found that the best results (F1 for 'value' class is 0.77 and F1 macro is 0.83) were achieved with SVM classification, which utilized embeddings from a fine-tuned BERT-model.

There are some limitations of our approach mostly pertaining to its generalizability. While our study employed machine learning and deep learning for classification, it is worth noting fair criticism regarding this approach. As has been noted, results of machine learning classification depend on the specific texts used for training the algorithm and thus are not universally generalizable (Ponizovskiy *et al.*, 2020). However, if social media data are representative and the machine learning approach uses not just words, but word-embeddings and employs transfer learning (that is, using pre-trained large models which can be fine-tuned on one's data), the results *can* be generalized to the entire social network.

It is not clear if the suggested approach and model would work in other social media, as the VKontakte audience differs in terms of, for instance, its education, orientation towards international integration, and oppositional political views (Koltsova *et al.*, 2021; Enikolopov *et al.*, 2018; Reuter & Szakonyi, 2015). Thus, to extend the applicability of the approach to other social networks, it might be necessary to supplement the model with network-specific data.

Despite these limitations, studying values in Russian social networks is of particular interest in the context of the Russian-Ukrainian conflict, and our proposed model is capable of providing value-expressive content for such analyses. Overall, our contribution should be regarded as a successful attempt to examine the value expression of social media posts from a reader's perspective.

[12] Testing beyond this study showed that ChatGPT has knowledge of the value categories in Schwartz's theory and can "understand" the meaning of a text in terms of a particular value category. This might imply that conducting more specific classification tasks by ChatGPT will yield better results.

## APPENDIX 1

### Text-annotation task

Since Yandex Toloka does not have a technical capability to filter out workers by their responses, we added the first line in the instruction: "This project is for regular users of the VKontakte network".

To ensure high quality of labeling, we have limited access to the annotation to the best 30% qualified coders (in terms of their internal ratings) and used their responses only if they were successful in completion of the training tasks (at least 15 correct annotations out of 20 posts with known codes). The crowd annotators saw the posts in batches of 10, each with one control task (that is a post with a clear value-expressive content). After incorrect labeling of at least two control posts, the annotator was banned. Some annotators provided responses too quickly (less than ## seconds per post) raising doubt in them paying attention to their work, therefore they were banned as well.

Since the posts were of different lengths (mean=35.5 words, min=1, max=2563) and therefore took different time to label, we uploaded batches of short (less than or equal to 35 words) and long posts (more than 35 words) separately, setting different payment for their labeling.

### Instruction:

*This project is for regular users of the VKontakte network!*

*You are taking part in a scientific study that aims to learn how to automatically recognize if any expression of life values is contained in social media posts.*

*Your task is to determine whether there is an expression of value in the publication that will be shown to you. In other words, you will be answering the question:*

*Does this post reflect the importance of any idea, phenomenon, or quality?*

*Yes, it reflects*

*Values are a person's beliefs about the importance of some object, phenomenon, or quality. Examples of values are:*

- *The importance of romantic relationships, love, fidelity, friendship, loyalty to loved ones (including love for children)*
- *The importance of equality of all people or the preservation of nature*
- *Desire for stability and order, following rules, importance of safety, need to protect one's health.*
- *Importance of being independent, desire to try new things, to take risks*
- *Importance of achieving goals, receiving deserved rewards*
- *Importance of simply enjoying life*

*Please keep in mind that other types of values may also appear in the texts - take into account your own understanding!*

*Also, keep in mind that text that reflects a value may:*

- *not speak directly about a particular value*
- *express a negation of a value (such texts should also be labeled as containing values)*

*Examples of value-reflective tests:*

- *"People are always criticizing. Don't let them break you. Achieving your goal is the best form of revenge" is an example of implicitly pointing out the importance of achieving one's goals.*

- *"The law of meanness is the only law that applies in our country!" - an example of an implicit indication of the importance of following the rules.*
- *"This really is one of the best days of our lives! Thank you, my dear" is an example of reflecting the value of love.*
- *"Sometimes you need to treat yourself #cakes #caféfroyo #eating" is an example of reflecting the value of pleasures.*

*No, it doesn't reflect*

*The text may not reflect the importance of an object or value. For example:*

- *Duty greetings, recipes (culinary or other), without personal evaluation*
- *Emotional statements that do not indicate the importance or value of something*

*Examples of texts that do not reflect values:*

- *"How touching! I even cried!" or "ay lovely 😂" - the subject of the statement is unclear.*
- *"Thank you all for the congratulations!" or "new year 2015" - Duty thank you and congratulations.*

*Unclear*

*There are texts that are difficult to understand due to the use of unfamiliar names, words, places, unclear context.*

*Spam*

*Texts written for advertising purposes, including on behalf of organizations. Types of spam:*

- *advertisements for the sale of goods and services*
- *request to subscribe to someone*
- *request to click on a link*
- *and other promotional posts, including those with veiled advertisements*

*Examples of spam texts:*

- *"open sale on direct flight Perm-Min.Vody 226-57-06" - an obvious announcement of sale*
- *"IF YOU WANT TO GO FORWARD, YOU HAVE TO LEARN!!!!". And you can do it with energetic people with positive vibes who are madly in love with their business and ready to share their knowledge and experience ... <...> #depilationorel #studyingthebest " - hidden advertising.*

**Statistics on crowd-worker annotations of the training dataset**

Table 6 – Training dataset annotation statistics by crowd-workers in Yandex Toloka

| Unique crowd-worker, number | 304 |
|---|---|
| Number of posts labelled by one crowd-worker, mean value | 56 |
| Time to label 10 posts, mean value (short posts, long posts) | 54 sec,<br>123 sec |
| Time in labeling, mean value (short posts, long posts) | 5 min,<br>11.5 min |

Thus, 304 crowd-workers participated in the labeling, on average each of them has labeled 56 posts, spending an average of 5 minutes labeling short posts and 11.5 minutes labeling long posts[13]. Therefore, speaking about the three labeling versions carried out by crowd-workers, we should keep in mind that we are dealing with not three, but with, in our case, 304 different perceptions. That is, we can talk about a kind of "collective intelligence" in relation to the resulting labeling that is received.

## APPENDIX 2

Table 7 – Translation of posts from Russian to English

| English | Russian |
|---|---|
| Do the ministers of economy and finance offer other options besides retirees? Or is this their only and last strategic reserve? | А кроме пенсионеров министры экономики и финансов предлагают другие возможности? Или у них это единственный и последний стратегический резерв? |
| Another one of my princesses has grown up! | Еще одна моя принцесса выросла |
| Well, do not feed bread, let me go somewhere! | Ну хлебом не корми, дай поехать куда-нибудь! |
| It's cool to snooze on the oceanfront wrapped in a towel🌊 # dominican#ocean#atlantic#ocean#palms🌴#vacation | Круто дремать на берегу океана, завернувшись в полотенце🌊 доминикана#океан#атлантическийокеан#пальмы🌴 #отпуск |
| There are no accidents – everything in this world is either a trial, a punishment, a reward, or a foreshadowing. ©Voltaire | Случайностей не существует — всё на этом свете либо испытание, либо наказание, либо награда, либо предвестие. ©Вольтер |
| If a guy loves you, he'll never let other people hurt you. He'll hurt you on his own | Если парень вас любит он никогда не даст вас в обиду. Он будет обижать вас сам |
| Instruction before the path of courage | Инструктаж перед тропой мужества)) |
| - Honey, at least start running in the morning... - If I run out, I won't come back!!! | -Дорогая, начни хотя бы бегать по утрам... -Если я выбегу,то уже не ВЕРНУСЬ!!! |
| Sometimes, I want to go far away from everyone, to a faraway place, not to see or even hear anyone. | Иногда, охота: Уеехать долеко от всех,в дальние края,чтобы никого не видеть, и даже. не слышать |
| The best psychologist is a friend with a bottle of champagne. | Самый лучший психолог - это подруга с бутылкой шампанского |
| Friends are everything to me (c) | Друзья всё для меня (с) |
| I'm finally an auntie!!!!!!!!)))))) | Я стала наконец-то тётей!!!!!!!!)))))) |
| Girls! 🙋‍♀️ What do you think the perfect makeup bag should consist of? 🤨 Of course, of those products that are just necessary to look perfect ☝ Cosmetics [Brand name removed] ✖ is just created for the perfect beauty kit. | Девочки! 🙋‍♀️ Как вы думаете, из чего должна состоять идеальная косметичка?🤨 Конечно, из тех средств, которые просто необходимы для того, чтобы выглядеть идеально☝ Косметика [Brand name removed] ✖ как раз создана для идеального бьюти-набора |

[13] Short posts accounted for 83% and long posts for 17% of the main dataset.

| | |
|---|---|
| Someday they will tell you that your kindness wipes their feet. Do not believe. Stay kind. After all, kindness is simple, and it will save the world. ☺ Sent via 💚 vk.com/ranks 💚 | Когда-нибудь тебе внушат, что о твою доброту вытирают ноги. Не верь. Оставайся добрым. Ведь добро — это просто, и оно спасает мир. ☺ Отправлено через 💚 vk.com/ranks 💚 |
| Tomorrow there is a place for a trial MASSAGE!!! TC Aksion Izhevsk | Завтра есть место на пробный МАССАЖ!!! Тц Аксион Ижевск |
| I remember! I'm proud! | Я помню! Я горжусь! |
| Sign the petition | Подпишите петицию |
| Christ is Risen! | Христос Воскресе |
| The secret world government, the Illuminati and the Zionists are after my personal data. I'm going underground | Тайное мировое правительство, иллюминаты и сионисты охотятся за моими личными данными. Я ухожу в подполье |
| Good morning! | Доброе утро! |
| Good night, everyone | Всем спок ночи |